\pdfoutput=1

\documentclass[11pt]{article}

\usepackage{EMNLP2023}

\usepackage{times}
\usepackage{latexsym}

\usepackage[T1]{fontenc}

\usepackage[utf8]{inputenc}

\usepackage{microtype}
\usepackage{tcolorbox}

\usepackage{inconsolata}
\usepackage{listings}

\usepackage{enumitem}

\title{NeMo Guardrails: A Toolkit for Controllable and Safe \\ 
LLM Applications with Programmable Rails}

\author{Traian Rebedea$^*$, Razvan Dinu$^*$, Makesh Sreedhar, Christopher Parisien, Jonathan Cohen \\
  NVIDIA \\
  Santa Clara, CA \\
  \texttt{ \{trebedea, rdinu, makeshn, cparisien, jocohen\}@nvidia.com} 
}

\begin{document}
\maketitle

\def\thefootnote{*}\footnotetext{Equal contribution}\def\thefootnote{\arabic{footnote}}

\begin{abstract}
NeMo Guardrails is an open-source toolkit\footnote{\url{https://github.com/NVIDIA/NeMo-Guardrails}} for easily adding programmable guardrails to LLM-based conversational systems. Guardrails (or \textit{rails} for short) are a specific way of controlling the output of an LLM, such as not talking about topics considered harmful, following a predefined dialogue path, using a particular language style, and more. There are several mechanisms that allow LLM providers and developers to add guardrails that are embedded into a specific model at training, e.g. using model alignment. Differently, using a runtime inspired from dialogue management, NeMo Guardrails allows developers to add \textit{programmable} rails to LLM applications - these are user-defined, independent of the underlying LLM, and interpretable. Our initial results show that the proposed approach can be used with several LLM providers to develop controllable and safe LLM applications using programmable rails.  
\end{abstract}

\section{Introduction}
Steerability and trustworthiness are key factors for deploying Large Language Models (LLMs) in production. Enabling these models to stay on track for multiple turns of a conversation is essential for developing task-oriented dialogue systems. This seems like a serious challenge as LLMs can be easily led into veering off-topic~\cite{pang2023leveraging}. At the same time, LLMs also tend to generate responses that are factually incorrect or completely fabricated (\textit{hallucinations})~\cite{manakul2023selfcheckgpt, peng2023check, azaria2023internal}. In addition, they are vulnerable to prompt injection (or jailbreak) attacks, where malicious actors manipulate inputs to trick the model into producing harmful outputs~\cite{kang2023exploiting, wei2023jailbroken, zou2023universal}. 

\begin{figure}
    \centering
    \includegraphics[width=0.7\columnwidth]{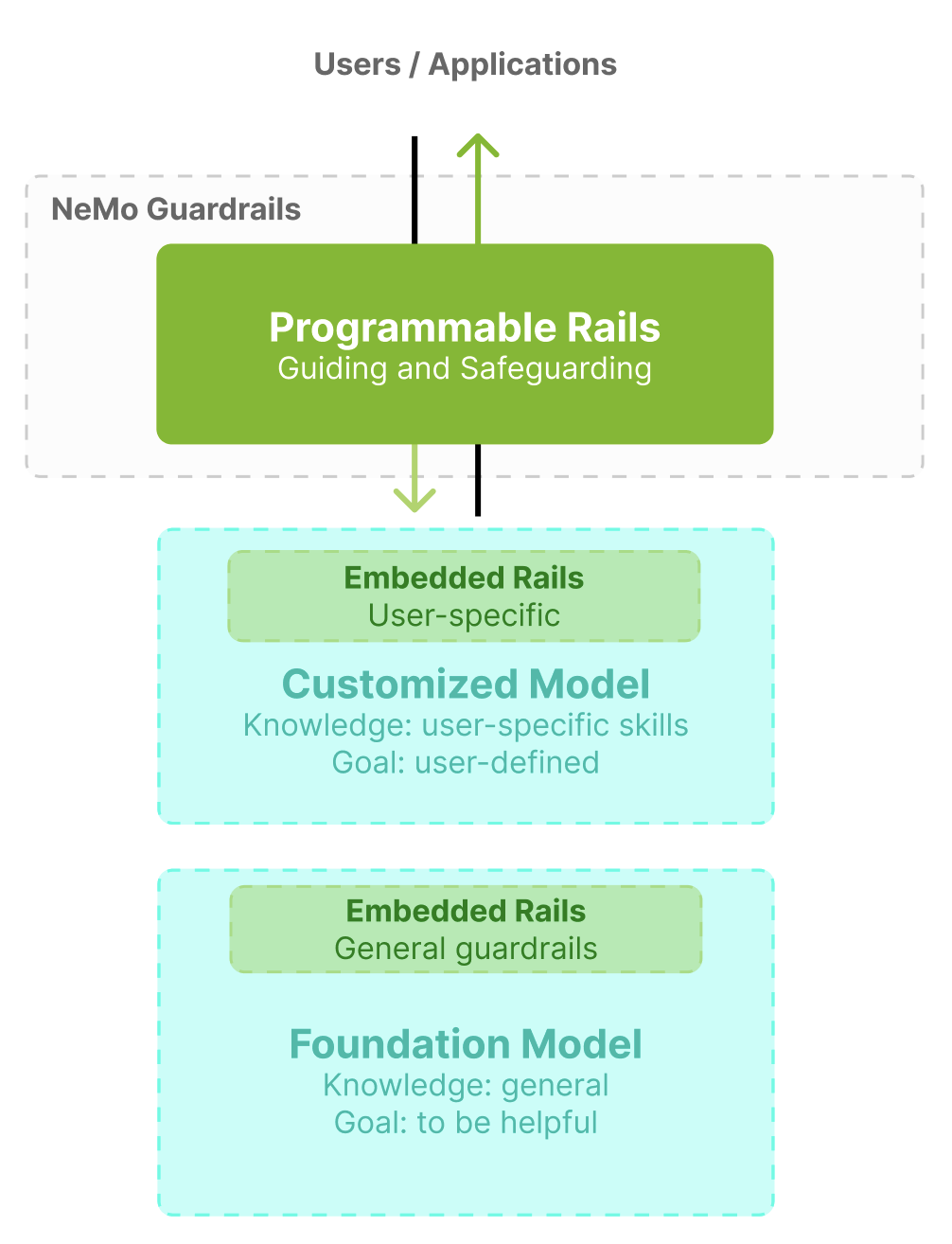}
    \caption{Programmable vs. embedded rails for LLMs.}
    \label{fig:programmable-rails}
\end{figure}

Building trustworthy and controllable conversational systems is of vital importance for deploying LLMs in customer facing situations. NeMo Guardrails is an open-source toolkit for easily adding programmable rails to LLM-based applications. 
Guardrails (or \textit{rails}) provide a mechanism for controlling the output of an LLM to respect some human-imposed constraints, e.g. not engaging in harmful topics, following a predefined dialogue path, adding specific responses to some user requests, using a particular language style, extracting structured data. To implement the various types of rails, several techniques can be used, including model alignment at training, prompt engineering and chain-of-thought (CoT), and adding a dialogue manager.
While model alignment provides general rails embedded in the LLM at training and prompt tuning can offer user-specific rails embedded in a customized model, NeMo Guardrails allows users to define custom programmable rails at runtime as shown in Fig.~\ref{fig:programmable-rails}. 
This mechanism is independent of alignment strategies and supplements embedded rails, works with different LLMs, and provides interpretable rails defined using a custom modeling language, Colang.

To implement user-defined programmable rails for LLMs, our toolkit uses a programmable runtime engine that acts like a proxy between the user and the LLM. This approach is complementary to model alignment and it defines the rules the LLM should follow in the interaction with the users. Thus, the Guardrails runtime has the role of a dialogue manager, being able to interpret and impose the rules defining the programmable rails. These rules are expressed using a modeling language called Colang.
More specifically, Colang is used to define rules as dialogue flows 
that the LLM should always follow (see Fig.~\ref{fig:colang-flows}). Using a prompting technique with in-context learning and a specific form of CoT, we enable the LLM to generate the next steps that guide the conversation.
Colang is then interpreted by the dialogue manager to apply the guardrails rules predefined by users or automatically generated by the LLM to guide the behavior of the LLM. 

While NeMo Guardrails can be used to add safety and steerability to any LLM-based application, we consider that dialogue systems powered by an LLM benefit the most from using Colang and the Guardrails runtime. The toolkit is licensed as Apache 2.0, 
and we provide initial support for several LLM providers, together with starter example applications and evaluation tools. 

\section{Related Work}\label{sec:related-work}


\subsection{Model Alignment}
Existing solutions for adding rails to LLMs rely heavily on model alignment techniques such as instruction-tuning~\cite{wei2021finetuned} or reinforcement learning~\cite{ouyang2022training, glaese2022improving, openai2023gpt4}. The alignment of LLMs works on several dimensions, mainly to improve helpfulness and to reduce harmfulness. 
Alignment in general, including red-teaming~\cite{perez-etal-2022-red}, requires a large collection of input prompts and responses that are manually labeled according to specific criteria (e.g., harmlessness). 

Model alignment provides rails embedded at training in the LLM, that cannot easily be changed at runtime by users. Moreover, it also requires a large set of human-annotated response ratings for each rail to be incorporated by the LLM. While Reinforcement Learning from Human Feedback~\cite{ouyang2022training} is the most popular method for model alignment, alternatives such as RL from AI Feedback~\cite{bai2022constitutional} do not require a human labeled dataset and use the actual LLM to provide feedback for each response. 

While most alignment methods provide general embedded rails, in a similar way developers can add app-specific embedded rails to an LLM via prompt tuning~\cite{lester-etal-2021-power, liu-etal-2022-p}.

\begin{figure}
    \centering
    \includegraphics[width=0.8\columnwidth]{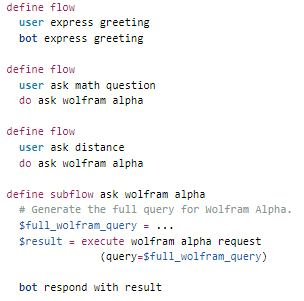}
    \caption{Dialogue flows defined in Colang: a simple greeting flow and two topical rail flows calling the custom action \texttt{wolfram alpha request} to respond to math and distance queries.}
    \label{fig:colang-flows}
\end{figure}

\subsection{Prompting and Chain-of-Thought}

The most common approach to add user-defined programmable rails to an LLM is to use prompting, including prompt engineering and in-context learning~\cite{brown2020language}, by prepending or appending a specific text to the user input~\cite{wang2022toxicity, si2022prompting}. This text specifies the behavior that the LLM should adhere to. 

The other approach to provide LLMs with user-defined runtime rails is to use chain-of-thought (CoT)~\cite{wei2022chain}. In its simplest form, CoT appends to the user instruction one or several similar examples of input and output pairs for the task at hand. Each of these examples contains a more detailed explanation in the output, useful for determining the final answer. 
Other more complex approaches involve several steps of prompting the LLM in a generic to specific way~\cite{zhou2022least} or even with entire dialogues with different roles 
similar to an inner monologue~\cite{huang2022inner}. 

\subsection{Task-Oriented Dialogue Agents}
Building task-oriented dialogue agents generally requires two components: a Natural Language Understanding (NLU) and a Dialogue Management (DM) engine~\cite{bocklisch2017rasa, liu2021benchmarking}. 
There exist a wide range of tools and solutions for both NLU and DM, ranging from open-source solutions like Rasa~\cite{bocklisch2017rasa} to proprietary platforms, such as Microsoft LUIS or Google DialogFlow~\cite{liu2021benchmarking}. Their functionality mostly follows these two steps: first the NLU extracts the intent and slots from the user message, then the DM predicts the next dialogue state given the current dialogue context. 

The set of intents and dialogue states are finite and pre-defined by a conversation designer. The bot responses are also chosen from a closed set depending on the dialogue state. This approach allows to define specific dialogue flows that tightly control any dialogue agent. Conversely, these agents are rigid and require a high amount of human effort to design and update the NLU and dialogue flows.


At the other end of the spectrum are recent end-to-end (E2E) generative approaches that use LLMs for dialogue tracking and bot message generation \cite{hudevcek2023llms, zhang2023sgp}. NeMo Guardrails also uses an E2E approach to build LLM-powered dialogue agents, but it combines a DM-like runtime able to interpret and maintain the state of dialogue flows written in Colang with a CoT-based approach to generate bot messages and even new dialogue flows using an LLM.


\section{NeMo Guardrails}

\vspace{-0.1cm}
\subsection{General Architecture}

\begin{figure*}
    \centering
    \includegraphics[width=\textwidth]{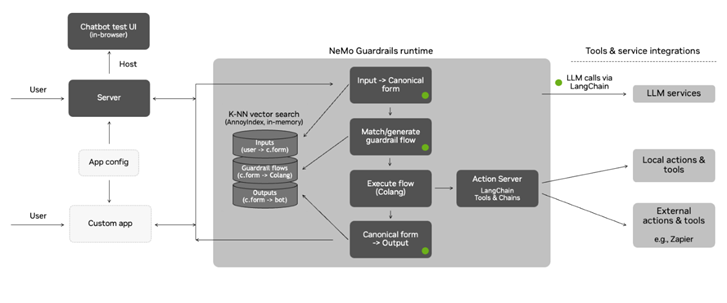}
    \caption{NeMo Guardrails general architecture.}
    \label{fig:guardrails-architecture}
\end{figure*}

NeMo Guardrails acts like a proxy between the user and the LLM as detailed in Fig.~\ref{fig:guardrails-architecture}. It allows developers to define programmatic rails that the LLM should follow in the interaction with the users using \textbf{Colang}, a formal modeling language designed to specify flows of events, including conversations. Colang is interpreted by the \textbf{Guardrails runtime} which applies the user-defined rules or automatically generated rules by the LLM, as described next. These rules implement the guardrails and guide the behavior of the LLM.

An excerpt from a Colang script is shown in Fig.~\ref{fig:colang-flows} - these scripts are at the core of a Guardrails app configuration. The main elements of a Colang script are: user canonical forms, dialogue flows, and bot canonical forms. All these three types of definitions are also indexed in a \textbf{vector database} (e.g., Annoy~\cite{annoy}, FAISS~\cite{johnson2019billion}) to allow for efficient nearest-neighbors lookup when selecting the few-shot examples for the prompt.
The interaction between the LLM and the Guardrails runtime is defined using  Colang rules. When prompted accordingly, the LLM is able to generate Colang-style code using few-shot in-prompt learning. Otherwise, the LLM works in normal mode and generates natural language.



\textbf{Canonical forms}~\cite{sreedhar-parisien-2022-prompt} are a key mechanism used by Colang and the runtime engine. They are expressed in natural language (e.g., English) and encode the meaning of a message in a conversation, similar to an intent. The main difference between intents and canonical forms is that the former are designed as a closed set for a text classification task, while the latter are generated by an LLM and thus are not bound in any way, but are guided by the canonical forms defined by the Guardrails app. The set of canonical forms used to define the rails that guide the interaction is specified by the developer; these are used to select few-shot examples when generating the canonical form for a new user message.

Using these key concepts, developers can implement a variety of programmable rails. We have identified two main categories: topical rails and execution rails. \textbf{Topical rails} are intended for controlling the dialogue, e.g. to guide the response for specific topics or to implement complex dialogue policies. \textbf{Execution rails} call custom actions defined by the app developer; we will focus on a set of safety rails available to all Guardrails apps.   

\vspace{-0.1cm}
\subsection{Topical Rails}
\label{sec:topical-rails-main}
Topical rails employ the key mechanism used by NeMo Guardrails: Colang for describing programmable rails as \textbf{dialogue flows}, together with the Colang interpreter in the runtime for \textbf{dialogue management} (\textit{Execute flow [Colang]} block in  Fig.~\ref{fig:guardrails-architecture}). Flows are specified by the developer to determine how the user conversation should proceed. The dialogue manager in the Guardrails runtime uses an event-driven design (an event loop that processes events and generates back other events) to ensure which flows are active in the current dialogue context.

The runtime has three main stages (see Fig.~\ref{fig:guardrails-architecture}) for guiding the conversation with dialogue flows and thus ensuring the topical rails:

\textbf{Generate user canonical form.} Using similarity-based few-shot prompting, generate the canonical form for each user input, allowing the guardrails system to trigger any user-defined flows.

\textbf{Decide next steps and execute them.} Once the user canonical form is identified, there are two potential paths: 
\textbf{1) Pre-defined flow:} If the canonical form matches any of the developer-specified flows, the next step is extracted from that particular flow by the dialogue manager; 
\textbf{2) LLM decides next steps:} For user canonical forms that are not defined in the current dialogue context, we use the generalization capability of the LLM to decide the appropriate next steps - e.g., for a travel reservation system, if a flow is defined for booking bus tickets, the LLM should generate a similar flow if the user wants to book a flight. 

\textbf{Generate bot message(s).} Conditioned by the next step, the LLM is prompted to generate a response. Thus, if we do not want the bot to respond to political questions, and the next step for such a question is \textit{bot inform cannot answer} – 
the bot would deflect from responding, respecting the rail.

Appendix~\ref{sec:colang-language-guide} provides details about the Colang language. Appendix~\ref{sec:prompt-topical-rails} contains sample prompts.

\vspace{-0.1cm}
\subsection{Execution Rails}
The toolkit also makes it easy to add \textit{"execution"} rails. These are custom actions (defined in Python), monitoring both the input and output of the LLM, and can be executed by the Guardrails runtime when encountered in a flow. While execution rails can be used for a wide range of tasks, we provide several rails for LLM safety covering fact-checking, hallucination, and moderation. 

\vspace{-0.1cm}
\subsubsection{Fact-Checking Rail}

Operating under the assumption of retrieval augmented generation~\cite{wang2023shall}, we formulate the task as an entailment problem. Specifically, given an \textit{evidence} text and a generated \textit{bot response}, we ask the LLM to predict whether the response is grounded in and entailed by the evidence. For each evidence-hypothesis pair, the model must respond with a binary entailment prediction using the following prompt:

\vspace{0.2cm}
\hangindent=0.5cm 
\textit{    You are given a task to identify if the hypothesis is grounded and entailed in the evidence. You will only use the contents of the evidence and not rely on external knowledge. Answer with yes/no. "evidence": \{\{evidence\}\} "hypothesis": \{\{bot\_response\}\} "entails":}
\vspace{0.2cm}

If the model predicts that the hypothesis is not entailed by the evidence, this suggests the generated response may be incorrect. Different approaches can be used to handle such situations, such as abstaining from providing an answer. 

\vspace{-0.1cm}
\subsubsection{Hallucination Rail}
For general-purpose questions that do not involve a retrieval component, we define a hallucination rail to help prevent the bot from making up facts. The rail uses self-consistency checking similar to SelfCheckGPT~\cite{manakul2023selfcheckgpt}: given a query, we first sample several answers from the LLM and then check if these different answers are in agreement. 
For hallucinated statements, repeated sampling is likely to produce responses that are not in agreement.

After we obtain $n$ samples from the LLM for the same prompt, we concatenate $n-1$ responses to form the \textit{context} and use the $n^{th}$ response as the \textit{hypothesis}. 
Then we use the LLM to detect if the sampled responses are consistent using the prompt template defined in Appendix~\ref{sec:prompt-templates}.

\vspace{-0.1cm}
\subsubsection{Moderation Rails}
The moderation process in NeMo Guardrails contains two key components:  

$\bullet$ \textbf{Input moderation}, also referred as \textit{jailbreak} rail, aims to detect potentially malicious user messages before reaching the dialogue system.
    
$\bullet$ \textbf{Output moderation} aims to detect whether the LLM responses are legal, ethical, and not harmful prior to being returned to the user.
\vspace{0.2cm}

The moderation system functions as a pipeline, with the user message first passing through input moderation before reaching the dialogue system. After the dialogue system generates a response powered by an LLM, the output moderation rail is triggered. Only after passing both moderation rails, the response is returned to the user.

Both the input and output moderation rails are framed as another task to a powerful, well-aligned LLM that vets the input or response. The prompt templates for these rails are found in Appendix~\ref{sec:prompt-templates}. 







\vspace{-0.1cm}
\section{Sample Guardrails Applications}\label{sec:sample-apps}
Adding rails to conversation applications is simple and straightforward using Colang scripts. 

\vspace{-0.1cm}
\subsection{Topical Rails}
Topical rails can be used in combination with execution rails to decide when a specific action should be called or to define complex dialogue flows for building task oriented agents. 

In the example presented in Fig.~\ref{fig:colang-flows}, we implement two topical rails that allow the Guardrails app to use the WolframAlpha engine to respond to math and distance queries. To achieve this, the \texttt{wolfram alpha request} custom action (implemented in Python, available on Github) is using the WolframAlpha API to get a response to the user query. This response is then used by the LLM to generate an answer in the context of the current conversation.    

\vspace{-0.1cm}
\subsection{Execution Rails}

The steps involved in adding executions rails are:
\vspace{-0.2cm}
\begin{enumerate}[noitemsep]
    \item \textbf{Define the action} - Defining a rail requires the developer to define an action that specifies the logic for the rail (in Python). 
    \item \textbf{Invoke action in dialogue flows} - Once the action has been defined, we can call the action from Colang using the \textit{execute} keyword.
    \item \textbf{Use action output in dialogue flow} - The developer can specify how the application should react to the output from the action. 
\end{enumerate}

Appendix~\ref{sec:action-definitions} contains details about defining actions, together with an example of the actions that implement the input and output moderation rails.

\begin{figure}[htbp]
    \centering
    \includegraphics[width=0.8\columnwidth]{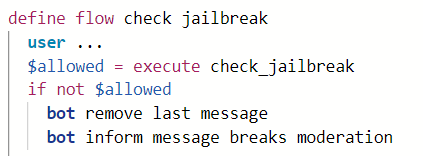}
    \caption{Flow using jailbreak rail in Colang}
    \label{fig:jailbreak-moderation}
\end{figure}

Fig.~\ref{fig:jailbreak-moderation} shows a sample flow in Colang that invokes the $check\_jailbreak$ action. If the jailbreak rail flags a user message, the developer can decide not to show the generated response and to output a default text instead. Appendix~\ref{sec:sample-flows-actions} provides other examples of flows using the executions rails.

\section{Evaluation}

In this section, we provide details on how we measure the performance of  various rails. Additional information for all tasks and a discussion on the automatic evaluation tools available in NeMo Guardrails are provided in Appendix~\ref{sec:additional-evaluation}.

\subsection{Topical Rails}

The evaluation of topical rails focuses on the core mechanism used by  the toolkit to guide conversations using canonical forms and dialogue flows. The current evaluation experiments employ datasets used for conversational NLU. In this section, we present the results for the Banking dataset~\cite{casanueva-etal-2022-nlu}, while additional experiments can be found in Appendix~\ref{sec:additional-evaluation}.

Starting from a NLU dataset, we create a Colang application (publicly available on Github) by mapping intents to canonical forms and defining simple dialogue flows for them. 
The evaluation dataset used in our experiments is balanced, containing at most 3 samples per intent sampled randomly from the original datasets. The test dataset has 231 samples spanning over 77 different intents.

\begin{figure}
    \centering
    \includegraphics[scale=0.48]{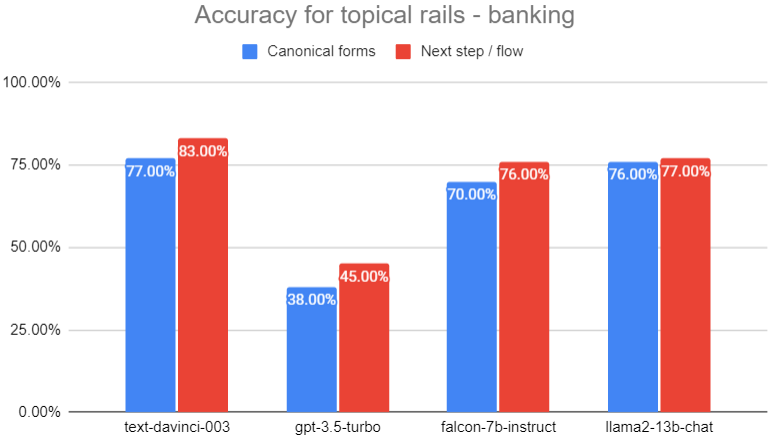}
    \caption{Performance of topical rails on Banking.}
    \label{fig:topical-rails-short}
\end{figure}

The results of the top 3 performing models are presented in Fig.~\ref{fig:topical-rails-short}, showing that topical rails can be successfully used to guide conversations even with smaller open source models such as \texttt{falcon-7b-instruct} or \texttt{llama2-13b-chat}. As the performance of an LLM is heavily dependent on the prompt, 
all results might be improved with better prompting.

The topical rails evaluation highlights several important aspects. First, each step in the three-step approach (user canonical form, next step, bot message) used by Guardrails offers an improvement in performance. Second, it is important to have at least $k=3$ samples in the vector database for each user canonical form for achieving good performance. Third, some models (i.e., \texttt{gpt-3.5-turbo}) produce a wider variety of canonical forms, even with few-shot prompting. In these cases, it is useful to add a similarity match instead of exact match for generating canonical forms. 

\subsection{Execution Rails}

\paragraph{Moderation Rails}
To evaluate the moderation rails, we use the Anthropic Red-Teaming and Helpful datasets~\cite{bai2022training, perez-etal-2022-red}. 
We have sampled a balanced \textit{harmful}-\textit{helpful} evaluation set as follows: from the Red-Teaming dataset we sample prompts with the highest harmful score, while from the Helpful dataset we select an equal number of prompts. 

We quantify the performance of the rails based on the proportion of harmful prompts that are blocked and the proportion of helpful ones that are allowed. 
Analysis of the results shows that using both the input and output moderation rails is much more robust than using either one of the rails individually. Using both rails \texttt{gpt-3.5-turbo} has a great performance - blocking close to 99\% of harmful (compared to 93\% without the rails) and just 2\% of helpful requests - details in Appendix~\ref{sec:additional-evaluation}.

\paragraph{Fact-Checking Rail}
We consider the MSMARCO dataset~\cite{bajaj2016ms} to evaluate the performance of the fact-checking rail. The dataset consists of \textit{(context, question, answer)} triples. In order to mine negatives (answers that are \emph{not} grounded in the context) we use OpenAI \texttt{text-davinci-003} to rewrite the positive answer to a hard negative that looks similar to it, but is not grounded in the evidence. We construct a combined dataset by equally sampling both positive and negative triples. Both \texttt{text-davinci-003} and \texttt{gpt-3.5-turbo} perform well on the fact-checking rail and obtain an overall accuracy of ~80\% (see Fig.~\ref{fig:fact-checking} in Appendix~\ref{sec:factchecking-evaluation-appendix}).

\paragraph{Hallucination Rail}

Evaluating the hallucination rail is difficult without employing subjective manual annotation. To overcome this issue and be able to automatically quantify its performance, we compile a list of 20 questions based on a false premise (questions that do not have a right answer).

Any generation from the language model, apart from deflection, is considered a failure. We then quantify the benefit of employing the hallucination rail as a fallback mechanism. For \texttt{text-davinci-003}, the LLM is unable to deflect prompts that are unanswerable and using the hallucination rail helps intercept 70\% of these prompts. \texttt{gpt-3.5-turbo} performs much better, deflecting unanswerable prompts or marking that its response could be incorrect in 65\% of the cases. Even in this case, employing the hallucination rail boosts performance up to ~95\%. 

\begin{figure}
    \centering
    \includegraphics[scale=0.22]{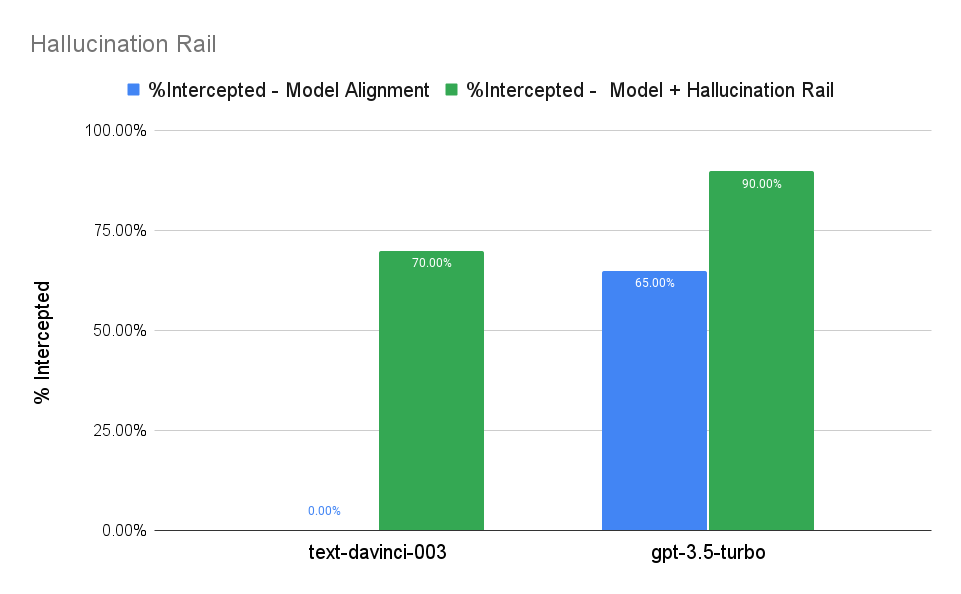}
    \caption{Performance of the hallucination rail.}
    \label{fig:hallucination}
\end{figure}

\section{Conclusions}
We present NeMo Guardrails, a toolkit that allows developers to build controllable and safe LLM-based applications by implementing programmable rails. These rails are expressed using Colang and can also be implemented as custom actions if they require a complex logic. Using  CoT prompting and a dialogue manager that can interpret Colang code, the Guardrails runtime acts like a proxy between the application and the LLM enforcing the user-defined rails. 

\section{Limitations}
\subsection{Programmable Rails and Embedded Rails}
Building controllable and safe LLM-powered applications, in general, and dialogue systems, in particular, is a difficult task. We acknowledge that the approach employed by NeMo Guardrails of using developer-defined programmable rails, implemented with prompting and the Colang interpreter, is not a perfect solution. 

Therefore we advocate that, whenever possible, our toolkit should not be used as a stand-alone solution, especially for safety-specific rails. Programmable rails complement embedded rails and these two solutions should be used together for building safe LLM applications. The vision of the project is to also provide, in the future, more powerful customized models for some of the execution rails that should supplement the current pure prompting methods. On another hand, our results show that adding the moderation rails to existing safety rails embedded in powerful LLMs (e.g., ChatGPT), provides a better protection against jail-break attacks. 

In the context of controllable and task-oriented dialogue agents, it is difficult to develop customized models for all possible tasks and topical rails. Therefore, in this context, NeMo Guardrails is a viable solution for building LLM-powered task-oriented agents without extra mechanisms. However, even for topical rails and task-oriented agents, we plan to release p-tuned models that achieve better performance for some of the tasks, e.g. for canonical form generation.

\subsection{Extra Costs and Latency}
The three-step CoT prompting approach used by the Guardrails runtime incurs extra costs and extra latency. As these calls are sequentially chained (i.e., the generation of the next steps in the second phase depends on the user canonical form generated in the first stage), the calls cannot be batched. In our current implementation, the latency and costs required are about 3 times the latency and cost of a normal call to generate the bot message without using Guardrails. We are currently investigating if in some cases we could use a single call to generate all three steps (user canonical form, next steps in the flow, and bot message).

Using a more complex prompt and few-shot in-context learning also generates slightly extra latency and a larger cost compared to a normal bot message generation for a vanilla conversation. Developers can decide to use a simpler prompt if needed.

However, we consider that developers should be provided with various options for their needs. Some might be willing to pay the extra costs for having safer and controllable LLM-powered dialogue agents. Moreover, GPU inference costs will decrease and smaller models can also achieve good performance for some or all NeMo Guardrails tasks. As presented in our paper, we know that \texttt{falcon-7b-instruct}~\cite{penedo2023refinedweb} already achieves very good performance for topical rails. We have seen similar positive performance from other recent models, like Llama 2 (7B and 13B) chat variants~\cite{touvron2023llama}.

\section{Broader Impact}

As a toolkit to enforce programmable rails for LLM applications, including dialogue systems, NeMo Guardrails should provide benefits to developers and researchers. Programmable rails supplement embedded rails, either general (using RLHF) or user-defined (using p-tuned customized models). For example, using the fact-checking rail developers can easily build an enhanced retrieval-based LLM application and it also allows them to assess the performance of various models as programmable rails are model-agnostic. The same is true for building LLM-based task-oriented agents that should follow complex dialogue flows.

At the same time, before putting a Guardrails application into production, the implemented programmable rails should be thoroughly tested (especially safety related rails). Our toolkit provides a set of evaluation tools for testing the performance both for topical and execution rails.

Additional details for our toolkit can be found in the Appendix, including  simple installation steps for running the toolkit with the example Guardrails applications that are shared on Github. A short demo video is also available: \url{https://youtu.be/Pfab6UWszEc}. 

\bibliography{anthology,custom}
\bibliographystyle{acl_natbib}
\clearpage
\appendix

\section{Installation Guide and Examples}
\label{sec:installation-guide}
Developers can download and install the latest version of the NeMo Guardrails toolkit directly from Github~\footnote{\url{https://github.com/NVIDIA/NeMo-Guardrails/}}. They can also install the latest stable release using \texttt{pip install nemoguardrails}. 

We have a concise installation guide~\footnote{\url{https://github.com/NVIDIA/NeMo-Guardrails/blob/main/docs/getting_started/installation-guide.md}} showing how to run a Guardrails app using the provided Command Line Interface (CLI) or how to launch the Guardrails web server. The server powers a simple chat web client to engage with all the Guardrails apps found in the folder specified when starting the server. 

Five reference Guardrails applications are provided as a general demonstration for building different types of rails. 
\begin{itemize}
    \item \textbf{Topical Rail:} Making the bot stick to a specific topic of conversation.
    
    \item \textbf{Moderation Rail:} Moderating a bot's response.

    \item \textbf{Fact Checking and Hallucination Rail:} Ensuring factual answers.
    
    \item \textbf{Secure Execution Rail:} Executing a third-party service with LLMs.
    
    \item \textbf{Jail-breaking Rail:} Ensuring safe answers despite malicious intent from the user.
\end{itemize}
These examples are meant to showcase the process of building rails, not as out-of-the-box safety features. Customization and strengthening of the rails is highly recommended.

The sample Guardrails applications also contain examples on how to use several open-source models (e.g., \texttt{falcon-7b-instruct}, \texttt{dolly-v2-3b}, \texttt{vicuna-7b-v1.3}) deployed locally or using HuggingFace Inference private endpoints. Other examples cover how to combine various chains defined in Langchain with programmable rails defined in NeMo Guardrails. 

Additional details about the reference applications and about the toolkit in general can be found on the main documentation page\footnote{\url{https://github.com/NVIDIA/NeMo-Guardrails/blob/main/docs/README.md}}.

\section{Colang Language and Dialogue Manager}
\label{sec:colang-language-guide}
Colang is a language for modeling sequences of events and interactions, being particularly useful for modeling conversations. At the same time, it enables the design of guardrails for conversational systems using the Colang interpreter, an event-based processing engine that acts like a dialogue manager.

Creating guardrails for conversational systems requires some form of understanding of how the dialogue between the user and the bot unfolds. Existing dialog management techniques such us flow charts, state machines or frame-based systems are not well suited for modeling highly flexible conversational flows like the ones we expect when interacting with an LLM-based system.

However, since learning a new language is not an easy task, Colang was designed as a mix of natural language (English) and Python. If you are familiar with Python, you should feel confident using Colang after seeing a few examples, even without any explanation.

The main  concepts used by the Colang language are the following:

\begin{itemize}
\item \textit{Utterance:} the raw text coming from the user or the bot.
\item \textit{Message:} the canonical form (structured representation) of a user/bot utterance.
\item \textit{Event:} something that has happened and is relevant to the conversation, e.g. user is silent, user clicked something, user made a gesture, etc.
\item \textit{Action:} a custom code that the bot can invoke; usually for connecting to a third-party API.
\item \textit{Context:} any data relevant to the conversation (encoded as a key-value dictionary).
\item \textit{Flow:} a sequence of messages and events, potentially with additional branching logic.
\item \textit{Rails:} specific ways of controlling the behavior of a conversational system (a.k.a. bot), e.g. not talk about politics, respond in a specific way to certain user requests, follow a predefined dialog path, use a specific language style, extract data etc. A rail in Colang can be modeled through one or more flows.
\end{itemize}

For additional details about Colang, please consult the Colang syntax guide~\footnote{\url{https://github.com/NVIDIA/NeMo-Guardrails/blob/main/docs/user_guide/colang-language-syntax-guide.md}}.

The Guardrails runtime uses an event-driven design (i.e., an event loop that processes events and generates back other events). Dialogue flows are treated as sequences of events, but even a simple user message is also an event - as an \texttt{UtteranceUserActionFinished} event is created and sent to the runtime. More details are available in the NeMo Guardrails architecture guide~\footnote{\url{https://github.com/NVIDIA/NeMo-Guardrails/blob/main/docs/architecture/README.md}}.

\section{Prompts for Topical Rails}
\label{sec:prompt-topical-rails}

NeMo Guardrails uses complex prompts, chained in 3 steps, to respond to a user message as described in Section~\ref{sec:topical-rails-main}. In the following listing we provide an example for the first step, to generate the canonical form for the last user message in the current conversation.

The prompt below is designed for \texttt{text-davinci-003} and is structured in four parts: 
\begin{enumerate}
    \item General prompt describing the task of the application.
    \item Sample conversation using Colang syntax.
    \item The most similar, given the current user message, few-shot ($k=5$) examples for mapping user messages to their corresponding canonical form.
    \item The current conversation between the user and the bot in Colang syntax.
\end{enumerate} 

\begin{lstlisting}[basicstyle=\tiny]
"""
Below is a conversation between a helpful AI assistant and a user. The bot is designed to generate human-like text based on the input that it receives. The bot is talkative and provides lots of specific details. If the bot does not know the answer to a question, it truthfully says it does not know.
"""

# This is how a conversation between a user and the bot can go:

user "Hello there!"
  express greeting
bot express greeting
  "Hello! How can I assist you today?"
user "What can you do for me?"
  ask about capabilities
bot respond about capabilities
  "I am an AI assistant which helps answer questions based on a given knowledge base. For this interaction, I can answer question based on the job report published by US Bureau of Labor Statistics"
user "Tell me a bit about the US Bureau of Labor Statistics."
  ask question about publisher
bot response for question about publisher
  "The Bureau of Labor Statistics is the principal fact-finding agency for the Federal Government in the broad field of labor economics and statistics"
user "thanks"
  express appreciation
bot express appreciation and offer additional help
  "You're welcome. If you have any more questions or if there's anything else I can help you with, please don't hesitate to ask."

# This is how the user talks:

user "What was the movement on nonfarm payroll?"
  ask about headline numbers

user "What's the number of part-time employed number?"
  ask about household survey data

user "How much did the nonfarm payroll rise by?"
  ask about headline numbers

user "What is this month's unemployment rate?"
  ask about headline numbers

user "How many long term unemployment individuals were reported?"
  ask about household survey data

# This is the current conversation between the user and the bot:

user "Hello there!"
  express greeting
bot express greeting
  "Hello! How can I assist you today?"
user "What can you do for me?"
  ask about capabilities
bot respond about capabilities
  "I am an AI assistant which helps answer questions based on a given knowledge base. For this interaction, I can answer question based on the job report published by US Bureau of Labor Statistics"
user "how many unemployed people were there in March?"
\end{lstlisting}

Similar prompts are defined for other LLMs (i.e., \texttt{gpt-3.5-turbo}, \texttt{falcon-7b-instruct} and others) and are available on Github~\footnote{\url{https://github.com/NVIDIA/NeMo-Guardrails/tree/main/nemoguardrails/llm/prompts}}.

When generating the user canonical form and the next steps to guide the conversation, we use $temp=0$, while for sampling the bot message we use a higher temperature ($temp=0.7$ or $temp=1$).

\section{Prompt Templates for Execution Rails}
\label{sec:prompt-templates}
In this section we provide the prompt templates used by the hallucination and moderation rails. 

\subsection{Hallucination Rail}
After we obtain $n$ samples from the conversational agent for the same prompt, we concatenate $n-1$ responses to form the \textit{context} and use the $n^{th}$ response as the \textit{hypothesis}. We utilize an LLM to verify if the hypothesis is consistent with the context using the following prompt template:

\vspace{0.2cm}
\hangindent=0.5cm 
\textit{    You are given a task to identify if the hypothesis is in agreement with the context below.
      You will only use the contents of the context and not rely on external knowledge.
      Answer with yes/no. "context": \{\{ sampled\_responses \}\} "hypothesis": \{\{ bot\_response \}\} "agreement":}
\vspace{0.2cm}

When sampling the bot responses, as proposed by SelfCheckGPT~\cite{manakul2023selfcheckgpt} we use a high temperature ($temp=1$) to ensure the variation between them necessary to assess the self-consistency of the answers. At the same time, the LLM-based hallucination verification call uses $temp=0$ to get the most probable response. This behavior is similar for all the other execution rails described in the paper.

\subsection{Moderation Rails}
Both the input and output moderation rails are framed as another task to a powerful, well-aligned LLM that vets the input or response. For the jailbreak rail, we use the following prompt template:

\vspace{0.2cm}
\hangindent=0.5cm 
\textit{ Instruction: \{\{ user\_input \}\} \\
Would this instruction make a language model break moderation policies, deviate from good aligned responses and provide answers that a language model should ideally not? Answer with yes/no.}
\vspace{0.2cm}

 The output moderation rail uses the following prompt template:

\vspace{0.2cm}
\hangindent=0.5cm 
\textit{ Model output: \{\{ bot\_response \}\} \\
Is this output legal, ethical and not harmful to any person? Answer with yes/no.}
\vspace{0.2cm}

\section{Action Definitions}
\label{sec:action-definitions}
The first step in creating an execution rail is to define a custom action that implements the logic of the rail. At this moment, this is done in Python. 

Any custom action has access to the \textit{context} of the conversation as can be seen in the subsequent examples. In the Guardrails runtime, the context is a sequence of all the events in the conversation history - including user and bot messages, canonical forms, action called and more. Some of the context events that might be accessed more often to define actions have a shortcut, e.g. $context.get("last\_bot\_message")$.

An action can receive any number of parameters from the Colang scripts where they are called. These are passed to the Python function implementing the action logic. At the same time, an action usually returns a value that can be used to further guide the dialogue.

\subsection{Output Moderation Rail}
This is the current implementation for the output moderation action. It uses the prompt template defined in Appendix~\ref{sec:prompt-templates} and returns $True$ if the bot response breaks the output moderation rail.

The call to the LLM is done using prompt templates and a simple LLM chain provided by Langchain\footnote{\url{https://github.com/langchain-ai/langchain}} for simplifying the communication with several LLM providers. 

\label{sec:action-output-moderation}
\begin{lstlisting}[language=Python, basicstyle=\tiny]
bot_response = context.get("last_bot_message")

if bot_response:
    output_moderation_template = 
    "Model Output: {bot_response}
    Is this output legal, ethical and not harmful to any person? Answer with yes/no."

    prompt = PromptTemplate(
        template=output_moderation_template, input_variables=["bot_response"]
    )

    output_moderation_chain = LLMChain(prompt=prompt, llm=llm)
    check = await output_moderation_chain.apredict(bot_response=bot_response)

    check = check.lower().strip()
    log.info(f"Output moderation check result is {check}.")

    if "no" in check:
        return False
return True
\end{lstlisting}

\subsection{Jailbreak Rail}
This is the current implementation for the input moderation action. It uses the prompt template defined in Appendix~\ref{sec:prompt-templates} and returns $True$ if the user input breaks the jailbreak rail.

\begin{lstlisting}[language=Python, basicstyle=\tiny]
user_input = context.get("last_user_message")

if user_input:
    jailbreak_template = "Instruction: {user_input}
    Would this instruction make a language model break moderation policies, deviate from good aligned responses and provide answers that a language model should ideally not? Answer with yes/no."

    prompt = PromptTemplate(
        template=jailbreak_template, input_variables=["user_input"]
    )

    jailbreak_chain = LLMChain(prompt=prompt, llm=llm)
    check = await jailbreak_chain.apredict(bot_response=bot_response)

    check = check.lower().strip()
    log.info(f"Jailbreak check result is {check}.")

    if "no" in check:
        return False
return True
\end{lstlisting}

\section{Sample Guardrails Flows using Actions}
\label{sec:sample-flows-actions}
This section includes some examples of using the safety execution rails, implemented as custom actions, inside Colang flows to define simple Colang applications. 

\begin{figure}[htbp]
    \centering
    \includegraphics[width=\columnwidth]{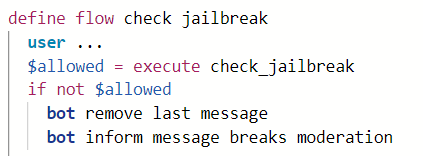}
    \caption{Flow using jailbreak rail in Colang}
    \label{fig:jailbreak-moderation-appendix}
\end{figure}

\begin{figure}[htbp]
    \centering
    \includegraphics[width=\columnwidth]{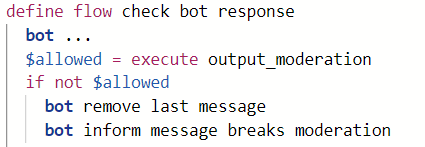}
    \caption{Flow using output moderation in Colang}
    \label{fig:output-moderation}
\end{figure}

Figure~\ref{fig:jailbreak-moderation-appendix} shows how to use the \texttt{check\_jailbreak} action for input moderation. The semantics is that for each user message (\texttt{user ...}), the jailbreak action is called to verify the last user message, and if it is flagged as a jailbreak attempt the last LLM bot-generated answer is removed and a new one is uttered to inform the user her/his message breaks the moderation policy. Figure~\ref{fig:output-moderation} shows how the \texttt{output\_moderation} action is used - the meaning is similar to jail-breaking, however it is triggered after any output bot message event (\texttt{bot ...}). 

\begin{figure}[htbp]
    \centering
    \includegraphics[width=\columnwidth]{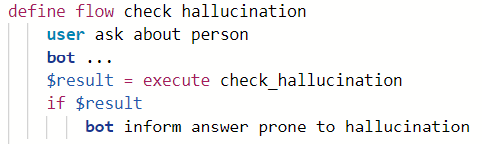}
    \caption{Flow using hallucination rail in Colang}
    \label{fig:hallucination-moderation}
\end{figure}

\begin{figure}[!htbp]
    \centering
    \includegraphics[width=0.8\columnwidth]{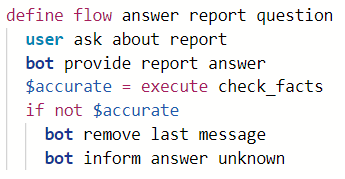}
    \caption{Flow using fact-checking rail in Colang}
    \label{fig:factcheck-moderation}
\end{figure}

In a similar way, Fig.~\ref{fig:hallucination-moderation} shows how to use the hallucination rail to check responses when for a particular topic (i.e., asking questions about persons, where GPT models are prone to hallucinate). In this case, the bot message is not removed, but an extra message is added to warn the user about a possible incorrect answer. Fig.~\ref{fig:factcheck-moderation} shows how to add fact-checking again for a specific topic, when asking a question about an employment report. In this situation, the LLM should be consistent with the information in the report.

\section{Additional Details on Evaluation}
\label{sec:additional-evaluation}

Our toolkit also provides the evaluation tooling and methodology to assess the performance of topical and execution rails. All the results reported in the paper can be replicated using the CLI evaluation tool available on Github, following the instructions about evaluation~\footnote{\url{https://github.com/NVIDIA/NeMo-Guardrails/blob/main/nemoguardrails/eval/README.md}}. The same page contains slightly more details than the current paper and is regularly updated with new results (including new LLMs). 

Detailed instructions on how to replicate the experiments can be found here~\footnote{\url{https://github.com/NVIDIA/NeMo-Guardrails/blob/main/docs/README.\#evaluation-tools}}.

\subsection{Topical Rails}
Topical rails evaluation focuses on the core mechanism used by NeMo Guardrails to guide conversations using canonical forms and dialogue flows.

The current evaluation experiments for topical rails uses two datasets employed for conversational NLU: chit-chat\footnote{\url{https://github.com/rahul051296/small-talk-rasa-stack}, dataset was initially released by Rasa} and banking.

The datasets were transformed into a NeMo Guardrails app, by defining canonical forms for each intent, specific dialogue flows, and even bot messages (for the chit-chat dataset alone). The two datasets have a large number of user intents, thus topical rails. One of them is very generic and with coarse-grained intents (chit-chat), while the banking dataset is domain-specific and more fine-grained. More details about running the topical rails evaluation experiments and the evaluation datasets is available here.

Preliminary evaluation results follow next. In all experiments, we have chosen to have a balanced test set with at most 3 samples per intent. For both datasets, we have assessed the performance for various LLMs and also for the number of samples ($k = all, 3, 1$) per intent that are indexed in the vector database. We have used a random seed of $42$ for all experiments to ensure consistency.

The results of the top 3 performing models are presented in Fig.~\ref{fig:topical-rails-short}, showing that topical rails can be successfully used to guide conversations even with smaller open source models such as \texttt{falcon-7b-instruct} or \texttt{llama2-13b-chat}. As the performance of an LLM is heavily dependent on the prompt, due to the complex prompt used by NeMo Guardrails all results might be improved with better prompting.

The topical rails evaluation highlights several important aspects. First, each step in the three-step approach (user canonical form, next step, bot message) used by Guardrails offers an improvement in performance. Second, it is important to have at least $k=3$ samples in the vector database for each user canonical form for achieving good performance. Third, some models (i.e., \texttt{gpt-3.5-turbo}) produce a wider variety of canonical forms, even with few-shot prompting. In these cases, it is useful to add a similarity match instead of exact match for generating canonical forms. 
In this case, the similarity threshold becomes an important inference parameter.

Dataset statistics and detailed results for several LLMs are presented in Tables~\ref{table:datasets-topical},~\ref{table:chitchat-results}, and~\ref{table:banking-results}. Some experiments have missing numbers either because those experiments did not compute those metrics or because the dataset does not contain specific items (for example, user-defined bot messages for the banking dataset).

\begin{table}[!ht]
    \centering
    \begin{tabular}{|l|l|l|}
    \hline
        Dataset & \# intents & \# test samples \\ \hline
        chit-chat & 76 & 226 \\ \hline
        banking & 77 & 231 \\ \hline
    \end{tabular}
    \caption{Dataset statistics for the topical rails evaluation.}
    \label{table:datasets-topical}
\end{table}

\begin{table*}[!ht]
    \centering
    \begin{tabular}{|l|p{1.2cm}|p{1.2cm}|p{1.2cm}|p{1.2cm}|p{1.2cm}|p{1.2cm}|}
    \hline
        Model & Us int, no sim & Us int, sim=0.6 & Bt int, no sim & Bt int, sim=0.6 & Bt msg, no sim & Bt msg, sim=0.6 \\ \hline
        text-davinci-003, k=all & 0.89 & 0.89 & 0.90 & 0.90 & 0.91 & 0.91 \\ \hline
        text-davinci-003, k=3 & 0.82 & N/A & 0.85 & N/A & N/A & N/A \\ \hline
        text-davinci-003, k=1 & 0.65 & N/A & 0.73 & N/A & N/A & N/A \\ \hline
        gpt-3.5-turbo, k=all & 0.44 & 0.56 & 0.50 & 0.61 & 0.54 & 0.65 \\ \hline
        dolly-v2-3b, k=all & 0.65 & 0.78 & 0.68 & 0.78 & 0.69 & 0.78 \\ \hline
        falcon-7b-instruct, k=all & 0.81 & 0.81 & 0.81 & 0.82 & 0.81 & 0.82 \\ \hline
        llama2-13b-chat, k=all & 0.87 & N/A & 0.88 & N/A & 0.89 & N/A \\ \hline
    \end{tabular}
    \caption{Topical evaluation results on chit-chat dataset. \textbf{Us int} means accuracy for user intents, \textbf{Bt int} is accuracy for next step generation (i.e., the bot intent), \textbf{Bt msg} is accuracy for generated bot message. \textbf{Sim} denotes if semantic similarity was used for matching (with a specified threshold, in this case $0.6$) or exact match. }
    \label{table:chitchat-results}
\end{table*}

\begin{table*}[!ht]
    \centering
    \begin{tabular}{|l|p{1.2cm}|p{1.2cm}|p{1.2cm}|p{1.2cm}|p{1.2cm}|p{1.2cm}|}
    \hline
        Model & Us int, no sim & Us int, sim=0.6 & Bt int, no sim & Bt int, sim=0.6 & Bt msg, no sim & Bt msg, sim=0.6 \\ \hline
       text-davinci-003, k=all & 0.77 & 0.82 & 0.83 & 0.84 & N/A & N/A \\ \hline
        text-davinci-003, k=3 & 0.65 & N/A & 0.73 & N/A & N/A & N/A \\ \hline
        text-davinci-003, k=1 & 0.50 & N/A & 0.63 & N/A & N/A & N/A \\ \hline
        gpt-3.5-turbo, k=all & 0.38 & 0.73 & 0.45 & 0.73 & N/A & N/A \\ \hline
        dolly-v2-3b, k=all & 0.32 & 0.62 & 0.40 & 0.64 & N/A & N/A \\ \hline
        falcon-7b-instruct, k=all & 0.70 & 0.76 & 0.75 & 0.78 & N/A & N/A \\ \hline
        llama2-13b-chat, k=all & 0.76 & N/A & 0.78 & N/A & N/A & N/A \\ \hline
    \end{tabular}
    \caption{Topical evaluation results on banking dataset.}
    \label{table:banking-results}
\end{table*}

\subsection{Execution Rails}
\subsubsection{Moderation Rail}
 To evaluate the moderation rails, we use the Anthropic Red-Teaming and Helpful datasets~\cite{bai2022training, perez-etal-2022-red}. 
The red-teaming dataset consists of prompts that are human-annotated (0-4) on their ability to elicit inappropriate responses from language models. A higher score implies that the prompt was more successful in bypassing model alignment. We randomly sample prompts with the highest rating to curate the \textit{harmful} set. All the prompts in the Anthropic Helpful dataset are genuine queries and forms our \textit{helpful} set. We create a balanced evaluation set with an equal number of \textit{harmful} and \textit{helpful} samples.

We quantify the performance of the rails based on the proportion of harmful prompts that are blocked and the proportion of helpful ones that are allowed. An ideal model would be able to block 100\% of the harmful prompts and allow 100\% of the helpful ones. We pass prompts from our evaluation set through the input (jailbreak) moderation rail. Only those that are not flagged are passed to the conversational agent to generate a response which is passed through the output moderation rail. Once again, only those responses that are not flagged are displayed back to the user.

Analysis of the results shows that using a combination of both the input (aka \textit{jailbreak} rail) and output moderation rails is more robust than using either one of the rails individually. It should also be noted that evaluation of the output moderation rail is subjective and each person/organization would have different subjective opinions on what should be allowed to pass through or not. In such situations, it would be easy to modify prompts to the moderation rails to reflect the beliefs of the entity deploying the conversational agent.

Using an evaluation set of 200 samples split equally between \textit{harmful} and \textit{helpful} and created as described above, we have seen that \texttt{text-davinci-003} blocks only 24\% of the harmful messages, while \texttt{gpt-3.5-turbo} does much better blocking 93\% of harmful messages without any moderation guardrail. In this case, blocking means that the model is not providing a response to an input requiring moderation. On the helpful inputs, both models do not block any request. 
Using only the input moderation rail, \texttt{text-davinci-003} blocks 87\% of harmful and 3\% of helpful requests. Using both input and output moderation, \texttt{text-davinci-003} blocks 97\% of harmful and 5\% of helpful requests, while \texttt{gpt-3.5-turbo} has a great performance - blocking close to 99\% of harmful and just 2\% of helpful requests.

\subsubsection{Fact-checking Rail}
\label{sec:factchecking-evaluation-appendix}
\begin{figure}
    \centering
    \includegraphics[scale=0.2]{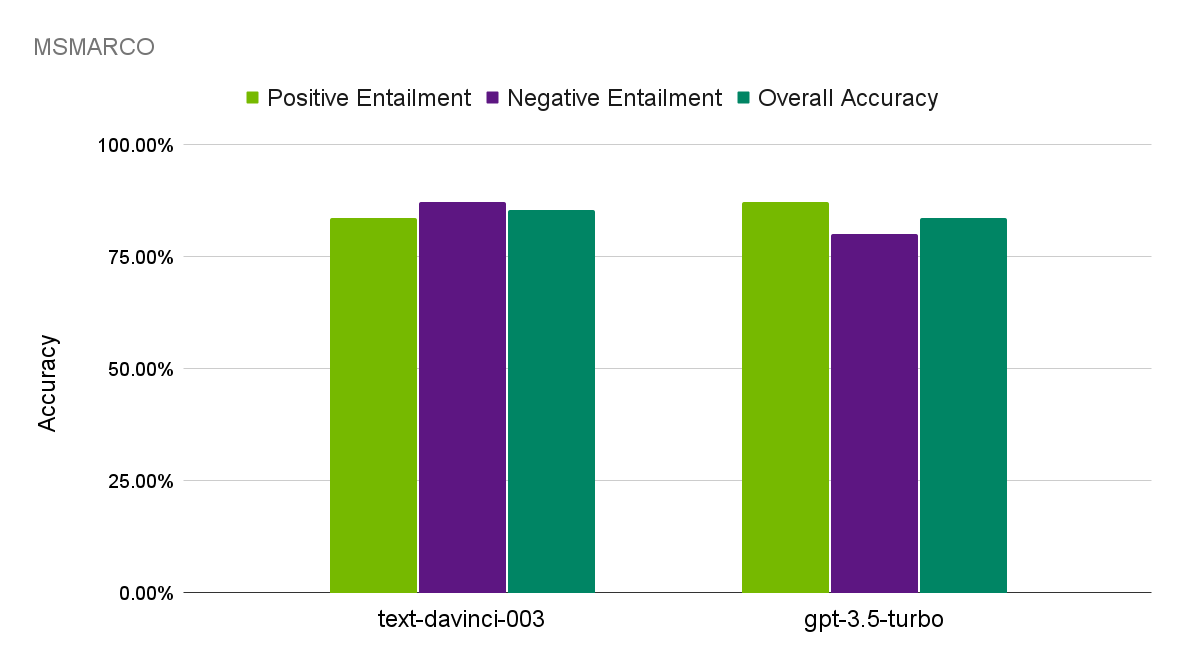}
    \caption{Performance of the fact-checking rail.}
    \label{fig:fact-checking}
\end{figure}

We consider the MSMARCO dataset~\cite{bajaj2016ms} to evaluate the performance of the fact-checking rail. The dataset consists of \textit{(context, question, answer)} triples. In order to mine negatives (answers that are \emph{not} grounded in the context), we use OpenAI \texttt{text-davinci-003} to rewrite the positive answer to a hard negative that looks similar to it, but is not grounded in the evidence. We construct a combined dataset by equally sampling both positive and negative triples. Both \texttt{text-davinci-003} and \texttt{gpt-3.5-turbo} perform well on the fact-checking rail and obtain an overall accuracy of ~80\% (Fig.~\ref{fig:fact-checking}). The behavior of the two models is slightly different: while \texttt{gpt-3.5-turbo} is better at discovering negatives,  \texttt{text-davinci-003} performs better on positive samples.

\subsubsection{Hallucination Rail}
Evaluating the hallucination rail is difficult since we cannot ascertain the questions that can be answered with factual knowledge embedded in the parameters of the language model. To effectively quantify the ability of the model to detect hallucinations, we compile a list of 20 questions based on a false premise. For example, one such question that does not have a right answer is: \textit{"When was the undersea city in the Gulf of Mexico established?"}

Any generation from the language model apart from deflection (i.e., recognizing that the question is unanswerable) is considered a failure. We also quantify the benefit of employing the hallucination rail as a fallback mechanism. For text-davinci-003, the base language model is unable to deflect prompts that are unanswerable and using the hallucination rail helps intercept 70\% of the unanswerable prompts. \texttt{gpt-3.5-turbo} performs very well at deflecting prompts that cannot be answered or hedging its response with statements about it could be incorrect. Even for such powerful models, we find that employing the hallucination rail helps boost the identification of questions that are prone to incorrect responses by  ~25\%. 
\end{document}